\documentclass[cameraready]{Interspeech}
\usepackage{comment}
\usepackage{subcaption}
\usepackage{xcolor, colortbl}
\usepackage{algorithm}
\usepackage{algpseudocode}
\usepackage{xspace}
\usepackage{multirow}
\usepackage{xurl}
\usepackage{tikz}
\usepackage{cite}
\usepackage{graphicx}
\usepackage{amssymb}
\usepackage{pifont}
\newcommand{\cmark}{\ding{51}}%
\newcommand{\xmark}{\ding{55}}%
\title{YODAS v3: Over 1 Million Hours of \\High-Bandwidth, Stereophonic, Multilingual Speech}

\author[affiliation={1}, equalcontribution]{William}{Chen}
\author[affiliation={2}, equalcontribution]{Shinnosuke}{Takamichi}
\author[affiliation={3}]{Sayaka}{Shiota}
\author[affiliation={4}]{Satoru}{Fukayama}
\author[affiliation={1}]{\\Samuele}{Cornell}
\author[affiliation={1}]{Shinji}{Watanabe}

\address{
    $^1$ Carnegie Mellon University, USA, $^2$ Keio University, Japan \\
    $^3$ Tokyo Metropolitan University, Japan \\
    $^4$ National Institute of Advanced Industrial Science and Technology (AIST), Japan
}

\email{williamchen@cmu.edu, shinnosuke\_takamichi@keio.jp, swatanab@andrew.cmu.edu}


\begin{document}

\maketitle

\begin{abstract}
    We present YODAS v3, a weakly-labeled speech corpus containing over 1.1 million hours\footnote{After acceptance, we identified and removed duplicated videos, reducing the total from 1.4M to 1.1M hours. The distribution figures and counts reflect the corrected dataset; the paper's findings are unchanged.} of 48kHz multi-channel audio in 147 languages, released under a CC BY 3.0 license. YODAS v3 is not only the largest open speech dataset to date, but also the first truly large-scale speech corpus with high-fidelity stereo audio. We first provide the collection methodology for the corpus, where we introduce new techniques for gathering language-balanced speech data. The effectiveness of our approach is shown by the language distribution of the crawled data: 22 languages in YODAS v3 have over 10K hours and 73 languages have over 5K hours of data. We then conduct extensive analyses on the composition of the data, such as the distribution of languages, audio quality, and transcription quality. Finally, we train baseline speech recognition and neural codec models to show the effectiveness of the dataset. Download at  \url{https://huggingface.co/datasets/espnet/yodas3}.
\end{abstract}

\setlength{\textfloatsep}{6pt}
\section{Introduction}

The rapid advancement of speech foundation models \cite{whisper, opuslmv, asru23-owsm, moshi, chen-etal-2024-towards-robust,pratap2023scaling,chu2024qwen2, chen2025owls, tian2026bagpiper, chen2026audiochat, barrault2023seamless, barrault2023seamlessm4t, omnilingual2025omnilingual, shi2025sam} is primarily driven by the availability of massive and diverse audio corpora \cite{voxpopuli, gigaspeech, yodas, kahnLibriLight, he2024emilia}. While proprietary systems benefit from internal datasets that can scale to over 10 million hours \cite{google-usm, comanici2025gemini} of audio, previous work \cite{chen-etal-2024-towards-robust} has shown that the combination of nearly all open speech corpora only sums to a little over 1 million hours of audio as of late 2024\footnote{While this number has now doubled to roughly 2 million hours after the release of \cite{luger-etal-2025-building}, it is still a far cry from that of propriety corpora.}.

Furthermore, there is also a quality gap between open and propriety corpora. The vast majority of open-source speech datasets, such as YODAS v2 \cite{yodas}, VoxPopuli \cite{voxpopuli}, MLS \cite{pratap2020mls}, and LibriLight \cite{kahnLibriLight} distribute data as monaural audio at sampling rates of 16 kHz or 24 kHz. While sufficient for standard ASR tasks, these constraints severely limit research in emerging domains that require high-resolution spatial audio, such as noisy multi-speaker ASR \cite{chime78}, high-fidelity audio codec modeling \cite{mousavi2025discrete}, full-band expressive TTS \cite{speakeasy}, spatial audio processing \cite{zheng2024bat}, and stereo speech enhancement \cite{li2025less}. Training models for these tasks requires high-resolution, multi-channel data, which has historically been scarce in open-access repositories.

Similar to how the release of corpora like VoxPopuli \cite{voxpopuli} and YODAS \cite{yodas} helped enable new areas of research in open large-scale self-supervised learning \cite{chen-etal-2024-towards-robust, pratap2023scaling, babu2021xls, boito2024mhubert} and large-scale supervised learning \cite{opuslmv, chen2025owls, peng25c_interspeech, sekoyan2025canary}, respectively, our goal is to empower future research that requires high-fidelty and/or spatial audio. We therefore introduce YODAS v3, a dataset containing more than one million hours of multilingual speech, making it the largest open speech dataset. Unlike prior large-scale collections, YODAS v3 preserves quality by saving it in the original downloaded format: 48kHz multi-channel OPUS files. Compared to other large-scale corpora, YODAS v3 is well-balanced across languages: 22 languages in YODAS v3 each have over 10K hours of data and 73 languages have over 5K hours. YODAS v3 also includes weak supervision in the form of automatically generated language identity tags, speech recognition transcripts, and English translations, making it also the first open supervised corpus at such scale. We first outline our data collection method for YODAS v3, in which we propose crawling techniques to better collect data for medium and low resource languages, and leads to a language-balanced corpora composition. We then perform analyses of the crawled data, including language distribution and audio characteristics. Finally, we train models using the crawled data to show the viability of our approach for Automatic Speech Recognition (ASR) and Neural Audio Codec modeling. 

\begin{table*}[t]
  \centering
    \caption{A comparison of YODAS v3 with a other open large-scale speech datasets. YODAS v3 is the first public dataset to reach a scale of over 1M hours while supporting high-bandwith stereo data.}
    \vspace{-0.1cm}
    \label{tab:data_comparison}
  \begin{tabular*}{\textwidth}{@{\extracolsep{\fill}}lrccccc}
    \toprule
    Dataset & Languages & Size & Sampling Rate & Stereo & Labeled & License \\
    \midrule
    Common Voice~\cite{commonvoice} & 137 & 0.033M hours & 48kHz & \xmark & \cmark & CC-0 \\
    MLS~\cite{pratap2020mls} & 8 & 0.051M hours & 16kHz & \xmark & \cmark & CC BY 4.0 \\
    FLEURS~\cite{FLEURS} & 102 & 0.001M hours & 16kHz & \xmark & \cmark & CC BY 2.5 \\
    VoxLingua107~\cite{voxlingua} & 107 &  0.007M hours & 16kHz & \xmark & \cmark & CC BY 4.0 \\
    Librilight~\cite{kahnLibriLight} & 1 & 0.060M hours & 16kHz & \xmark & \xmark & CC BY 4.0 \\
    VoxPopuli~\cite{voxpopuli} & 23 & 0.400M hours & 16kHz & \xmark & \xmark & CC-0\\
    Emilia~\cite{he2024emilia} & 6 & 0.100M hours & 24kHz & \xmark & \cmark & CC BY NC 4.0\\
    Unsupervised People's Speech~\cite{luger-etal-2025-building} & 89 & 0.740M hours & 48kHz & \cmark & \xmark & CC BY SA 4.0\\
    YODAS v2~\cite{yodas} & 140 & 0.550M hours & 24kHz & \xmark & \cmark & CC BY 3.0 \\
    \midrule
    YODAS v3 (this work) &  147 & 1.100M hours &  48kHz & \cmark & \cmark & CC BY 3.0 \\
    \bottomrule
  \end{tabular*}
\end{table*}

\setlength{\textfloatsep}{6pt}
\vspace{-0.1cm}
\section{Data Collection}
    \vspace{-0.1cm}
    \subsection{Keyword Generation} \label{sec:keyword}
    \vspace{-0.1cm}
        We first create lists of keywords for YouTube video search. In YODAS v2~\cite{yodas}, a shared keyword list across all languages is constructed from a multilingual Wikipedia dump data. This approach biases the keywords towards that of high-resource languages (such as English). This also results in reduced search quality, since words from the wrong language are used in the search term. Instead, we apply the following language-specific filters to create a keyword list for each language. For a given language, we download that language’s Wikipedia dump file and filter keywords that satisfy all of the following conditions:
        \begin{itemize}
            \item Not consisting of only punctuation and numbers.
            \item Neither punctuation nor a digit is the first character.
            \item 3--30 characters in length.
            \item Neither a filename, DOI (digital object identifier), a Wikipedia Template file, nor a Wikipedia Request file.
        \end{itemize}

        The final step is to filter out excessively rare words so that the keyword list is a manageable size for searching. We first train a unigram-based subword tokenizer \cite{kudo-2018-subword} on the filtered keywords for each language. We predefine a unigram vocabulary size: if we cannot construct a vocabulary at that size, we halve the size and try again until we successfully build a vocabulary. After training, we tokenize each keyword and compute its unigram likelihood. We finalized the keyword list for each language using only keywords with highest likelihood. 

    \vspace{-0.1cm}
    \subsection{Video Search}
        We search for videos IDs using the generated keyword list. Similar to YODAS v2, we limit the search to only videos that are uploaded under a Creative Commons license. We also prioritize more recently uploaded videos, since YouTube’s default search engine settings favor videos that are highly relevant to the keywords and therefore risks selecting only high-view-count videos. Prioritizing newer upload dates mitigates this issue and enables discovery of new videos.
        
        \begin{figure}[t]
            \centering
            \includegraphics[width=0.98\linewidth]{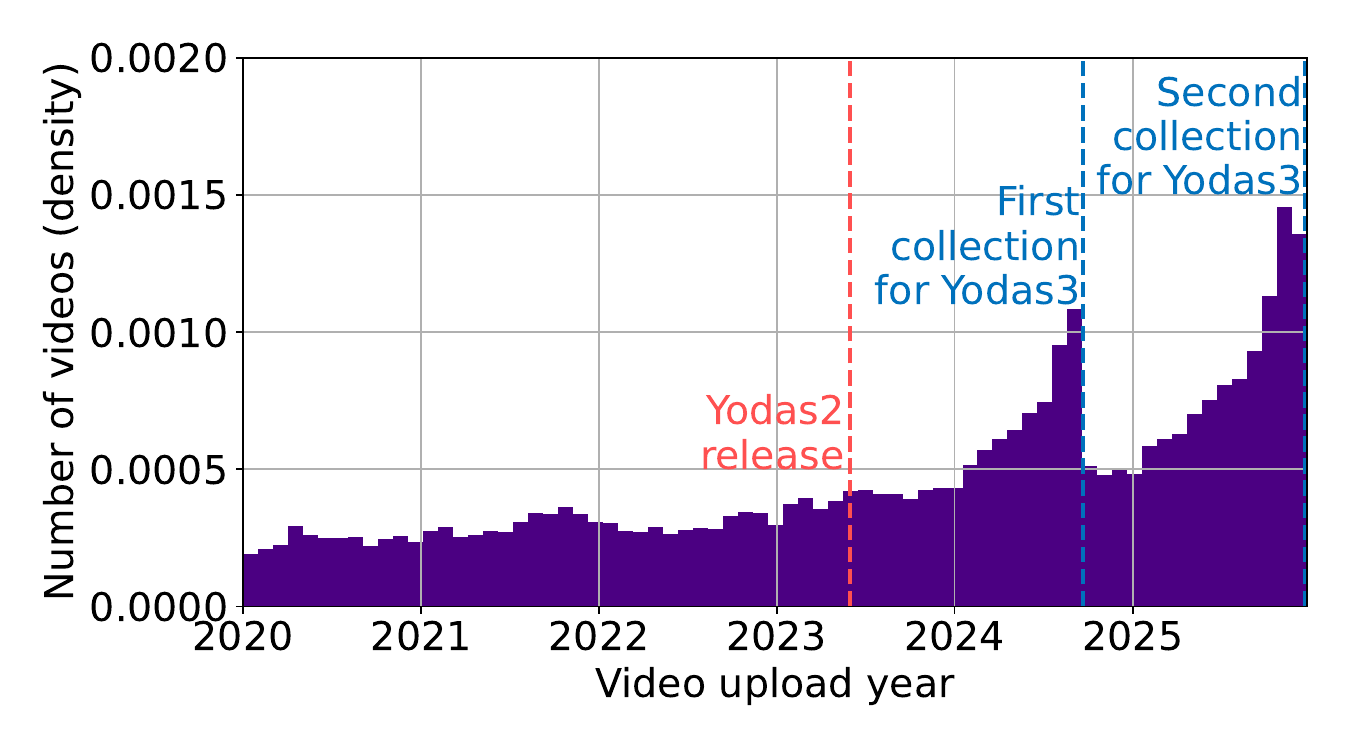}
            \vspace{-0.1cm}
            \caption{Histogram of video upload date. ``YODAS2'' is the submission deadline of ASRU 2023.}
            \label{fig:upload_date_distribution}
        \end{figure}

        Figure~\ref{fig:upload_date_distribution} shows the upload date histogram for the searched videos. Performing two video searches revealed that the setting effectively discovered newly uploaded videos not found in the previous search. To prevent data duplication, our final list of videos only contains IDs not found in YODAS v2, allowing both datasets to be combined in future work.

    \vspace{-0.1cm}
    \subsection{Downloading}
        We download the videos retrieved via the search and save their corresponding subtitles and audio. Audio is always downloaded and saved at the highest possible quality, which is 48kHz multi-channel OPUS. We always download any user-uploaded manual subtitles in the video's original language when they are available. Otherwise, we download the automatically generated YouTube subtitles in the original language. For non-English videos, we download the English subtitles as well, enabling future research on large-scale speech translation. All transcripts and translations are timestamped at the utterance level, allowing YODAS v3 to be used for both short-form and long-form tasks. Finally, we also provide the descriptions of each video, which can be used for summarization or retrieval tasks. In general, we observed that manual subtitles were quite rare - they amount to only 3.7\% of the total data (compared to 20\% in YODAS v2). We hypothesize that this is a consequence of the improvements in automatic subtitling quality since the crawling of YODAS v2, making the human production of manual subtitles less necessary. An overview of the meta-data included with each downloaded audio is shown in Table \ref{tab:metadata}.

\begin{table}[]
    \centering
    \caption{Examples of the metadata for each downloaded video}
    \label{tab:metadata}
    \begin{tabular}{l|l}
    \toprule
    Metadata     &  Example\\
    \midrule
    Channels & 2\\
    Effective channels & 2 \\
    Sampling rate & 48kHz\\
    Max bandwith & 20kHz\\
    Views & 67 \\
    Language & French\\
    Transcript & ''Bonjour mes amis..."\\
    Translation & ''Hello my friends..."\\
    Description & ''Une chanson pour..."\\
    \bottomrule
    \end{tabular}
\end{table}

\vspace{-0.1cm}
\section{Analysis}
\begin{figure*}
    \centering
    \includegraphics[width=1\linewidth]{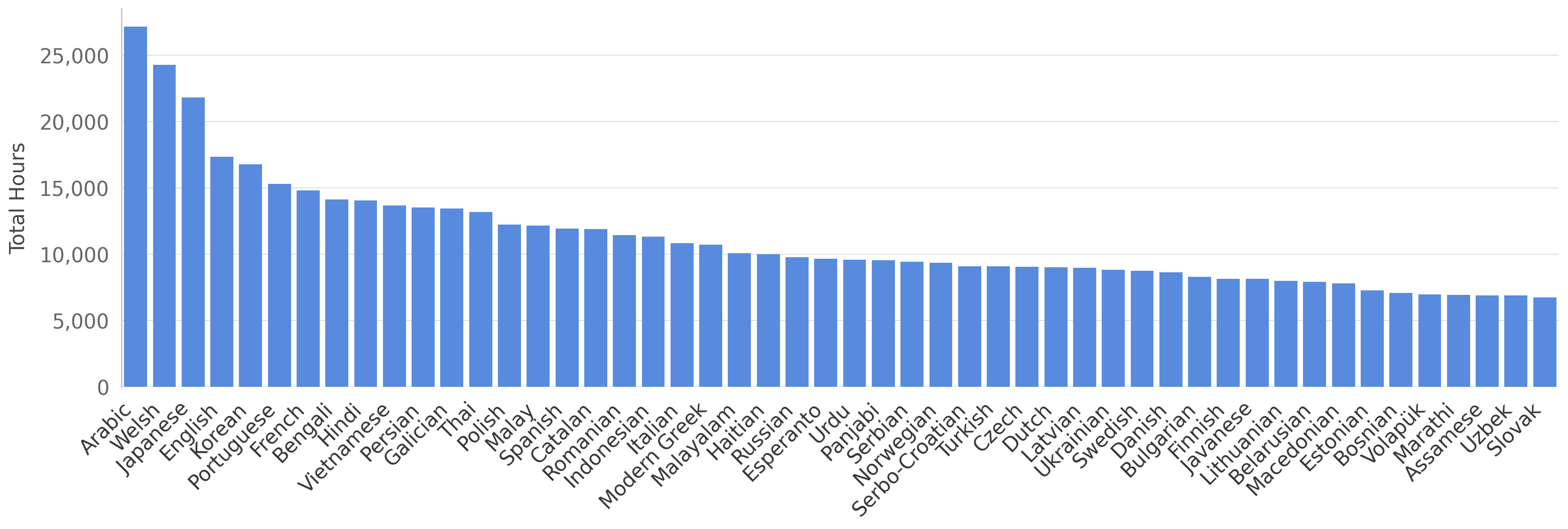}
    \vspace{-0.2cm}
    \caption{Data distribution of the top 50 languages in YODAS v3.}
    \label{fig:langs}
\end{figure*}
\subsection{Language Distribution}
A distribution of the data by language\footnote{We use the YouTube-provided locale of each video as a proxy for language identity due to the cost of automatic language ID at scale.} is shown in Figure \ref{fig:langs}.  The top 2 languages are Arabic and Welsh, each with roughly 25K hours of data. Japanese and English are the languages with the third and fourth amount of data at roughly 21K and 17K hours. Overall, 22 languages have over 10K hours of data and 73 languages have over 5K hours of data. This highlights the effectivness of our language-balanced video search approach (Section \ref{sec:keyword}): the mean and median amount of data per language is 7400 and 5180 hours, respectively. Compared to most datasets, where English is often more than half of the dataset \cite{yodas, pratap2020mls, commonvoice}, English is less than 2\% of YODAS v3. 

\subsection{Length Distribution}

\setlength{\textfloatsep}{6pt}
\begin{figure}
    \centering
    \includegraphics[width=\linewidth]{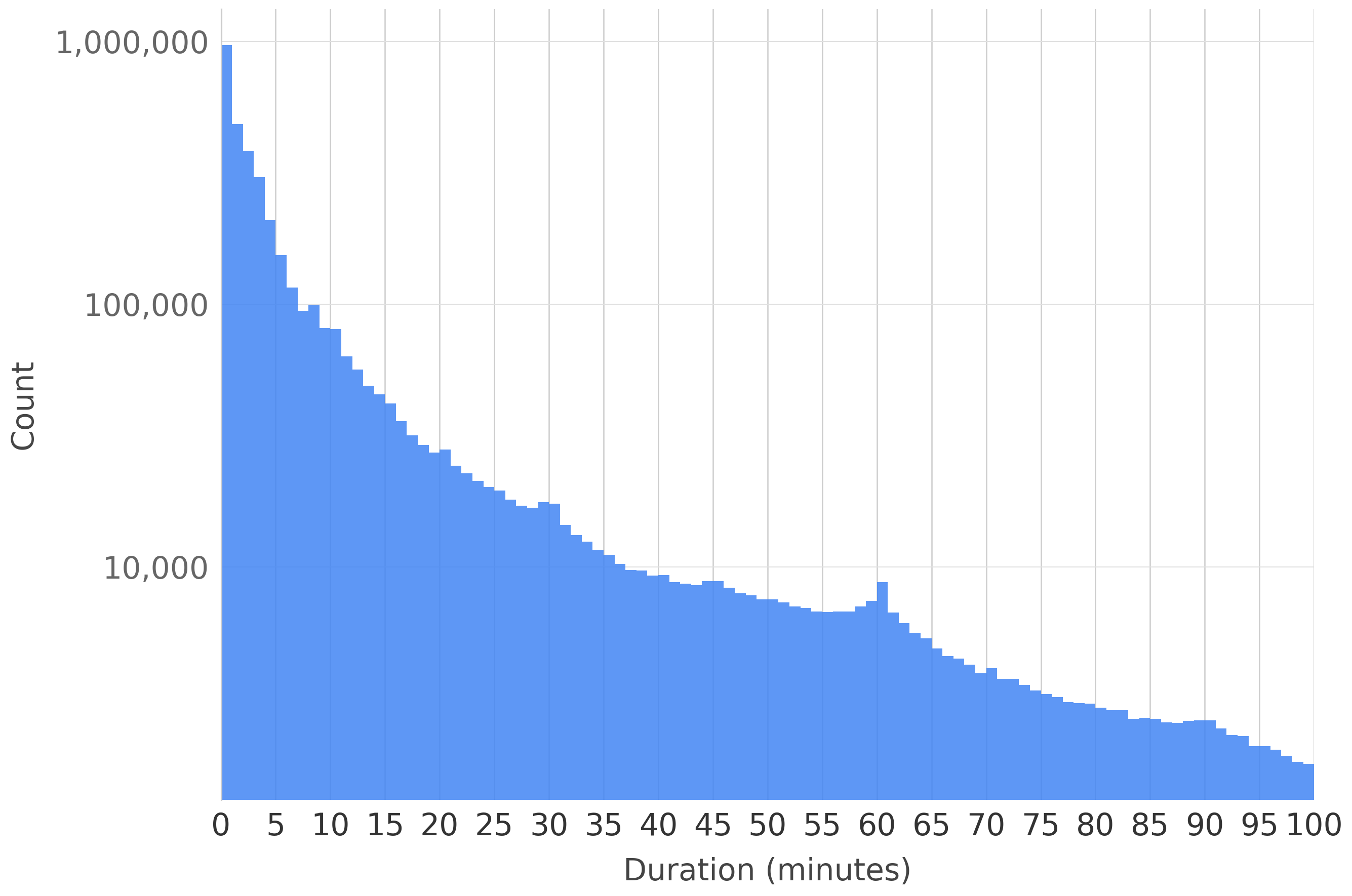}
    \vspace{-0.6cm}
    \caption{Distribution of videos by total length (log scale), bucketed into groups by minute.}
    \label{fig:length}
\end{figure}

Having ample long-form data is critical for frontier audio processing tasks like conversational ASR \cite{cornell23_chime}, speech summarization, and dialogue modeling. In this section, we analyze the lengths of the original crawled data. Figure \ref{fig:length} shows the distribution of lengths per video, bucketed into minute-level groups. Overall, 1.7 million videos are longer than 5 minutes (1 million hours total) and 1.2 million are longer than 10 minutes (950K hours total). Interestingly, we observe that the curve follows a long-tail distribution, with more than 24\% of the 4 million videos being less than one minute long. We attribute this observation to the rise in popularity of YouTube Shorts (short-form vertical videos similar to that of TikTok). Conversely, a manual qualitative analysis showed that a large amount of the longest videos (60+ minutes) are live-stream recordings. We leave further analysis these phenomena to future work.

\vspace{-0.1cm}
\subsection{True Bandwith Estimation}
\vspace{-0.1cm}

\begin{figure}[t]
    \centering
    \includegraphics[width=1\linewidth]{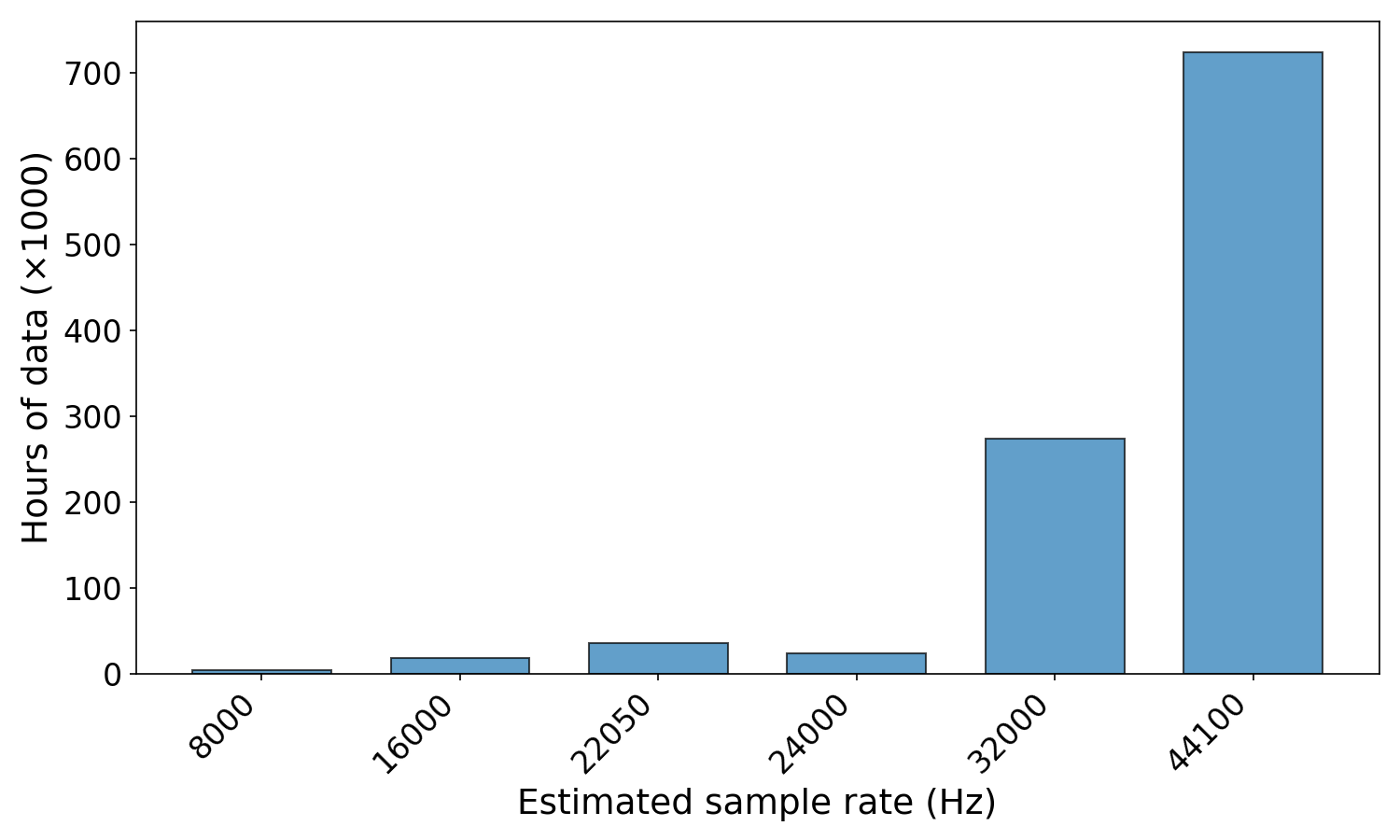}
    \vspace{-0.6cm}
    \caption{YODAS\,v3 data distribution by estimated effective bandwidth; over 92\% of recordings exceed 32\,kHz.}
    \label{fig:bandwith}
\end{figure}

We estimated the effective audio bandwidth of all recordings in the corpus, which are stored in WebM/Opus format. This step is necessary because the Opus codec always produces 48\,kHz PCM at decode time, regardless of its internal bandwidth mode (narrowband at 4\,kHz, wideband at 8\,kHz, up to fullband at most 20\,kHz). The nominal sample rate of decoded audio therefore provides no information about actual spectral content. Additionally, recordings may originate from limited-bandwidth capture devices (e.g., built-in laptop microphones, USB headsets) whose effective frequency response could fall well below their digitization rate.
For each file, we computed the short-time Fourier transform (STFT) with an 8192-sample Hann window and 50\% overlap. For each STFT frame, we identified the highest frequency bin whose magnitude exceeded 0.5\% of the frame's peak magnitude. Silent frames (root mean square (RMS) $\leq 10^{-6}$) were skipped. The estimated bandwidth for a recording is the maximum over all these analyzed frames. For recordings shorter than 5 minutes, the entire audio was decoded and analyzed. For longer recordings, five 60-second segments were drawn at random positions and the maximum bandwidth estimate across segments was retained.
This estimated maximum frequency was mapped to the smallest standard sample rate whose Nyquist frequency equals or exceeds it, from the set ${8,16,22.05,24,32,44.1}$\,kHz. This mirrors the classification used in the URGENT 2025 challenge~\cite{zhang25j_interspeech}.
The results are reported in Figure~\ref{fig:bandwith}. 
We can see that most of the data (67.3\%) has an effective bandwidth of 44.1\,kHz and 92.5\% of the data is above 32\,kHz. 

It should be noted that such bandwidth analysis is rarely conducted in prior dataset releases, despite being critical for understanding true audio fidelity. For instance, \cite{zhang25j_interspeech} found that CommonVoice and DNS5 LibriVox have effectively \textit{all} samples ($\sim$100\% each) with mismatched bandwidths, while even LibriTTS~\cite{zen2019libritts} is affected in $\sim$25\% of samples. Our analysis confirms that YODAS v3 is genuinely high-bandwidth, making it uniquely suited for tasks that require wideband audio.

\setlength{\textfloatsep}{6pt}
\vspace{-0.2cm}
\subsection{Audio Channels}
\vspace{-0.1cm}
\begin{figure}[t]
    \centering
    \includegraphics[width=1\linewidth]{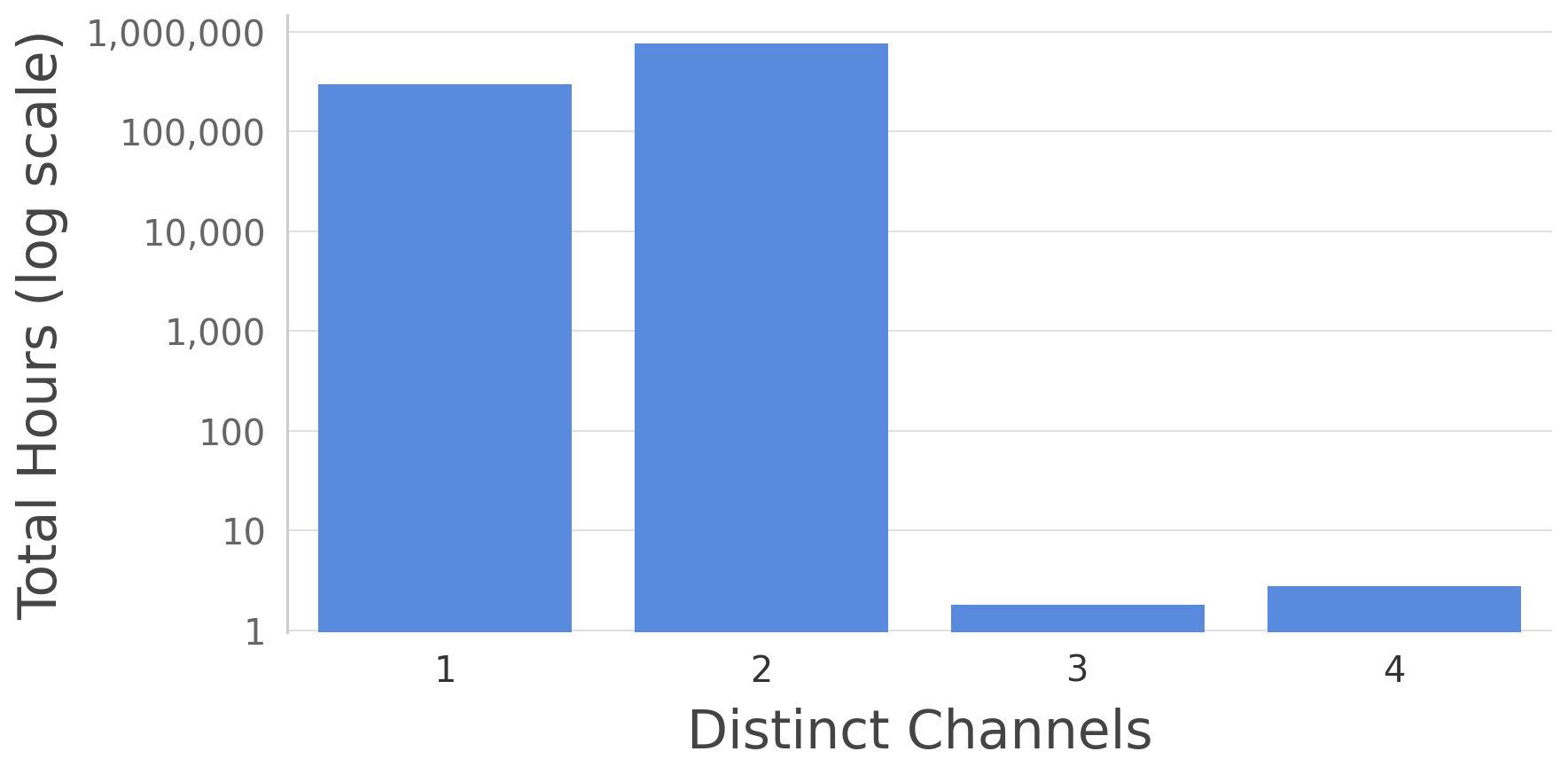}
    \caption{Log-scale distribution of data by effective number of channels in each audio file}
    \vspace{-0.1cm}
    \label{fig:channel}
\end{figure}

We also examine if the crawled data is truly multi-channel, or if the channels are in-fact duplicates of each other. For each pair of channels in an audio file, we subtract them from each other and compute the Root Mean Square (RMS) of the residual normalized by the average RMS of the two channels. We apply a threshold of 1e-3 to the result - values lower than the threshold indicate that the two compared channels are effectively identical. The distribution of results, in terms of the effective channel count, are shown in Figure \ref{fig:channel}. We find that over 71\% of all audio files (about 780K hours) are truly multi-channel, with the vast majority in being stereo data. We note that there is also a small amount of data that are effectively 3-4 channels - a result of the inclusion of 5.1 surround sound audio data. 

\vspace{-0.2cm}
\section{Experiments}
\vspace{-0.1cm}
\subsection{Speech Recognition} \label{sec:asr}

\noindent \textbf{Setup:} We adopt a similar experimental setup to \cite{yodas} for probing the quality of large-scale data by training monolingual ASR models on a random 7 language subset. We use 4500 hours for English and roughly 100 hours each for the other 6 languages. We process the data following the procedures proposed by OWSM v4 \cite{peng25c_interspeech}: CTC-segmentation, language-based filtering, and CTC score-based filtering. We train multiple models, each corresponding to different thresholds of the CTC score quantile filter: $\theta_{CTC}$ = 0.00 (no filtering), 0.10, 0.20, and 0.30. Models are intialized from OWSM v4 base 102M \cite{peng25c_interspeech} and trained for 40K steps using ESPnet \cite{espnet}. We evaluate the models on 7 languages from CommonVoice, using Word Error Rate (WER) for the 6 alphabet-based languages and Character Error Rate (CER) for Japanese.

\noindent \textbf{Results:} The evaluation results are shown in Table \ref{tab:asr}. Unlike prior work on filtering and cleaning YODAS v2 \cite{peng25c_interspeech}, which observed inconsistent results when balancing between data scale and filtering strictness, we find that using more data (lower $\theta_{CTC}$) with YODAS v3 leads to improved results. In fact, WER almost always improves for all 7 languages as $\theta_{CTC}$ decreases - the best models for each language (except for English) were trained without any score-based filtering, showing that YODAS v3 is directly usable out-of-the-box. This is a large improvement from YODAS v2, which required signficant filtering to avoid failed training runs (100+ WER). We hypothesize that this may be due to higher-quality automatic transcripts, which reflects the improvements in ASR performance since the release of YODAS v2 almost 3 years ago.

\begin{table}[tb]
    \centering
    \caption{WERs on different languages in Common Voice of ASR models trained on different data filtering thresholds.}
    \label{tab:asr}
    \resizebox{\columnwidth}{!}{
    \begin{tabular}{l|ccccccc}
    \toprule
    Threshold & Eng. & Deu. & Fra. & Jap. & Por. & Rus. & Vie.\\
    \midrule
    0.00 & 15.9 & \textbf{14.4} & \textbf{17.2} & \textbf{27.7} & \textbf{11.1} &  \textbf{15.3} & \textbf{20.6} \\
    0.10 & 15.3 & 14.6 & 17.8 & 29.2 & 11.5 & 15.4 & 21.1\\
    0.20 & 15.3 & 15.4 & 18.8 & 30.9 & 12.2 & 16.0 & 20.8\\
    0.30 & 15.6 & 16.3 & 21.8 & 34.5 & 13.4 & 16.7 & 23.1\\
    \bottomrule
    \end{tabular}}
\end{table}

\subsection{Neural Audio Codecs}
\noindent \textbf{Setup:} We train neural audio codecs on a random 1200 hour subset of the data at sample rates of 16kHz, 24kHz, and 48kHz. All models use the DAC architecture \cite{kumar2023high}. Models are trained using the official configurations from the ESPnet-Codec toolkit \cite{espnet_codec}, allowing for a fully fair comparison with models trained using existing datasets \cite{libritts, espnet_codec}. Specifically, we compare against models trained on LibriTTS \cite{libritts} (585 hours of English read speech)and AMUSE (33K hours of multilingual speech, music, and sound) \cite{espnet_codec}. We use the VERSA toolkit \cite{shi2024versa} for evaluation, measuring Short-Time Objective Intelligibility (STOI) \cite{stoi} and WavLM embedding-based speaker similarity \cite{ChenWavLm} on a 1000 video subset of YODAS v3 for in-domain testing and LibriTTS \cite{libritts} for out-of-domain evaluation.

\noindent \textbf{Results:} All models perform worse on the YODAS v3 test set compared to LibriTTS (Table \ref{tab:codec}). This is expected since the audio domain is much more challenging - clips often have background noise and multiple sound sources. However, LibriTTS contains only clean single speaker recordings. The best performing model is the 48kHz DAC trained on YODAS v3 on both the  in-domain and out-of-domain test sets with a STOI of 0.97 / 0.94 and a speaker similarity of 0.84 / 0.90, respectively. It outperforms all models trained on lower sampling rates and even models trained on more data (AMUSE, 33K hours), showing the effectiveness of the YODAS v3's high-bandwith data. 

\begin{table}[t]
    \centering
    \caption{Codec evaluation results. AMUSE is a multi-domain mixture for speech, sounds, and music, used by ESPnet-Codec.}
    \label{tab:codec}
    \begin{tabular}{lc|cc|cc}
    \toprule
    & & \multicolumn{2}{c}{LibriTTS} & \multicolumn{2}{c}{YODAS v3} \\
    Training Data & SR & STOI & SPK & STOI & SPK \\
    \midrule
    \textit{Baselines} \\
    LibriTTS & 16kHz & 0.95 & 0.76 & 0.74 & 0.69\\
    LibriTTS & 24kHz & \textbf{0.97} & 0.83 & 0.80 & 0.78\\
    AMUSE & 16kHz & 0.93 & 0.74 & 0.81 & 0.83 \\
    AMUSE & 44kHz & 0.95 & 0.82 & 0.90 & 0.77  \\
    \midrule
    \textit{This work} \\
    YODAS v3 & 16kHz & 0.92 & 0.60 & 0.82 & 0.79 \\
    YODAS v3 & 24kHz & 0.92 & 0.66 & 0.84 & 0.82 \\
    YODAS v3 & 48kHz & \textbf{0.97} & \textbf{0.84} & \textbf{0.94} & \textbf{0.90} \\
    \bottomrule
    \end{tabular}
\end{table}

\vspace{-0.1cm}
\section{Conclusion}
\vspace{-0.1cm}
This paper presents YODAS v3, a weakly-supervised corpus containing 1.1 million hours of high-bandwith stereo data across 147 languages.  YODAS v3 is created using a novel language-balanced keyword search approach, allowing us to collect large amounts of data for medium and low-resource languages: 22 languages in YODAS v3 have over 10K hours and 73
languages have over 5K hours of data. We conduct several analyses on the collected audio, showing how the data is both truly stereo and high-bandwith. Finally, we show the downstream use cases of YODAS v3. We train train ASR models are different subsets of the data, yielding results that show the cleainliness of the YODAS v3 transcripts. We also develop new high-fidelity neural codecs that are competitive with the state-of-the-art while showing how previous approaches fail to generalize to the complex audio scenes found in web-scale data.

\clearpage
\section{Acknowledgments}
Parts of this work used the PSC Bridges2 system and Delta/DeltaAI system at NCSA through allocations CIS210014 and IRI120008P from the ACCESS program, supported by NSF grants \#2138259,\#:2138286, \#:2138307, \#:2137603, and \#:2138296. This paper is also based on results obtained from a project, Programs for Bridging the gap between R\&D and the IDeal society (society 5.0) and Generating Economic and social value (BRIDGE)/Practical Global Research in the AI × Robotics Services, implemented by the Cabinet Office, Government of Japan.

\section{Generative AI Disclosure}
Generative AI was used to refine the manuscript text, such as wording and prose. It was also used to help write and scale the code necessary for large-scale crawling and data analyses. The final versions of all artificats produced with the assistance of generative AI were created and verified by humans.

\bibliographystyle{IEEEtran}
\bibliography{mybib}

\begin{thebibliography}{10}
\providecommand{\url}[1]{#1}
\csname url@samestyle\endcsname
\providecommand{\newblock}{\relax}
\providecommand{\bibinfo}[2]{#2}
\providecommand{\BIBentrySTDinterwordspacing}{\spaceskip=0pt\relax}
\providecommand{\BIBentryALTinterwordstretchfactor}{4}
\providecommand{\BIBentryALTinterwordspacing}{\spaceskip=\fontdimen2\font plus
\BIBentryALTinterwordstretchfactor\fontdimen3\font minus \fontdimen4\font\relax}
\providecommand{\BIBforeignlanguage}[2]{{%
\expandafter\ifx\csname l@#1\endcsname\relax
\typeout{** WARNING: IEEEtran.bst: No hyphenation pattern has been}%
\typeout{** loaded for the language `#1'. Using the pattern for}%
\typeout{** the default language instead.}%
\else
\language=\csname l@#1\endcsname
\fi
#2}}
\providecommand{\BIBdecl}{\relax}
\BIBdecl

\bibitem{whisper}
A.~Radford, J.~W. Kim, T.~Xu, G.~Brockman, C.~Mcleavey, and I.~Sutskever, ``Robust speech recognition via large-scale weak supervision,'' in \emph{ICML 2023}, 2023.

\bibitem{opuslmv}
J.~Tian, W.~Chen, Y.~Peng, J.~Shi, S.~Arora, S.~Bharadwaj, T.~Maekaku, Y.~Shinohara, K.~Goto, X.~Yue, H.~Yang, and S.~Watanabe, ``{OpusLM: A Family of Open Unified Speech Language Models},'' in \emph{{Interspeech 2025}}, 2025, pp. 3259--3263.

\bibitem{asru23-owsm}
Y.~Peng, J.~Tian, B.~Yan, D.~Berrebbi, X.~Chang, X.~Li, J.~Shi, S.~Arora, W.~Chen, R.~Sharma, W.~Zhang, Y.~Sudo, M.~Shakeel, J.~weon Jung, S.~Maiti, and S.~Watanabe, ``Reproducing {W}hisper-style training using an open-source toolkit and publicly available data,'' in \emph{ASRU 2023}, 2023.

\bibitem{moshi}
A.~D{\'e}fossez \emph{et~al.}, ``Moshi: a speech-text foundation model for real-time dialogue,'' \emph{arXiv preprint arXiv:2410.00037}, 2024.

\bibitem{chen-etal-2024-towards-robust}
\BIBentryALTinterwordspacing
W.~Chen, W.~Zhang, Y.~Peng, X.~Li, J.~Tian, J.~Shi, X.~Chang, S.~Maiti, K.~Livescu, and S.~Watanabe, ``Towards robust speech representation learning for thousands of languages,'' in \emph{Proceedings of the 2024 Conference on Empirical Methods in Natural Language Processing}, Y.~Al-Onaizan, M.~Bansal, and Y.-N. Chen, Eds.\hskip 1em plus 0.5em minus 0.4em\relax Miami, Florida, USA: Association for Computational Linguistics, Nov. 2024, pp. 10\,205--10\,224. [Online]. Available: \url{https://aclanthology.org/2024.emnlp-main.570/}
\BIBentrySTDinterwordspacing

\bibitem{pratap2023scaling}
V.~Pratap, A.~Tjandra, B.~Shi, P.~Tomasello, A.~Babu, S.~Kundu, A.~Elkahky, Z.~Ni, A.~Vyas, M.~Fazel-Zarandi \emph{et~al.}, ``Scaling speech technology to 1,000+ languages,'' \emph{arxiv:2305.13516}, 2023.

\bibitem{chu2024qwen2}
Y.~Chu, J.~Xu, Q.~Yang, H.~Wei, X.~Wei, Z.~Guo, Y.~Leng, Y.~Lv, J.~He, J.~Lin \emph{et~al.}, ``Qwen2-audio technical report,'' \emph{arXiv preprint arXiv:2407.10759}, 2024.

\bibitem{chen2025owls}
\BIBentryALTinterwordspacing
W.~Chen, J.~Tian, Y.~Peng, B.~Yan, C.-H.~H. Yang, and S.~Watanabe, ``{OWLS}: Scaling laws for multilingual speech recognition and translation models,'' in \emph{Forty-second International Conference on Machine Learning}, 2025. [Online]. Available: \url{https://openreview.net/forum?id=xnPW7yYomF}
\BIBentrySTDinterwordspacing

\bibitem{tian2026bagpiper}
J.~Tian, H.~Wang, B.-H. Su, C.-y. Huang, Q.~Wang, J.~Shi, W.~Chen, X.~Gong, S.~Arora, C.-J. Li \emph{et~al.}, ``Bagpiper: Solving open-ended audio tasks via rich captions,'' \emph{arXiv preprint arXiv:2602.05220}, 2026.

\bibitem{chen2026audiochat}
W.~Chen, P.~Seetharaman, R.~Kumar, O.~Nieto, S.~Watanabe, J.~Salamon, and Z.~Jin, ``Audiochat: Unified audio storytelling, editing, and understanding with transfusion forcing,'' \emph{arXiv preprint arXiv:2602.17097}, 2026.

\bibitem{barrault2023seamless}
L.~Barrault, Y.-A. Chung, M.~C. Meglioli, D.~Dale, N.~Dong, M.~Duppenthaler, P.-A. Duquenne, B.~Ellis, H.~Elsahar, J.~Haaheim \emph{et~al.}, ``Seamless: Multilingual expressive and streaming speech translation,'' \emph{arxiv:2312.05187}, 2023.

\bibitem{barrault2023seamlessm4t}
L.~Barrault, Y.-A. Chung, M.~C. Meglioli, D.~Dale, N.~Dong, P.-A. Duquenne, H.~Elsahar, H.~Gong, K.~Heffernan, J.~Hoffman \emph{et~al.}, ``{SeamlessM4T}-massively multilingual \& multimodal machine translation,'' \emph{arxiv:2308.11596}, 2023.

\bibitem{omnilingual2025omnilingual}
A.~Omnilingual, G.~Keren, A.~Kozhevnikov, Y.~Meng, C.~Ropers, M.~Setzler, S.~Wang, I.~Adebara, M.~Auli, C.~Balioglu \emph{et~al.}, ``Omnilingual asr: Open-source multilingual speech recognition for 1600+ languages,'' \emph{arXiv preprint arXiv:2511.09690}, 2025.

\bibitem{shi2025sam}
B.~Shi, A.~Tjandra, J.~Hoffman, H.~Wang, Y.-C. Wu, L.~Gao, J.~Richter, M.~Le, A.~Vyas, S.~Chen \emph{et~al.}, ``Sam audio: Segment anything in audio,'' \emph{arXiv preprint arXiv:2512.18099}, 2025.

\bibitem{voxpopuli}
C.~Wang \emph{et~al.}, ``{VoxPopuli: A Large-Scale Multilingual Speech Corpus for Representation Learning, Semi-Supervised Learning and Interpretation},'' in \emph{ACL 2021}, 2021.

\bibitem{gigaspeech}
G.~Chen \emph{et~al.}, ``{GigaSpeech}: An evolving, multi-domain {ASR} corpus with 10,000 hours of transcribed audio,'' in \emph{Interspeech 2021}, 2021.

\bibitem{yodas}
X.~Li, S.~Takamichi, T.~Saeki, W.~Chen, S.~Shiota, and S.~Watanabe, ``{YODAS}: {Y}outube-oriented dataset for audio and speech,'' in \emph{ASRU 2023}, 2023.

\bibitem{kahnLibriLight}
J.~Kahn, M.~Rivière, W.~Zheng, E.~Kharitonov, Q.~Xu, P.~Mazaré, J.~Karadayi, V.~Liptchinsky, R.~Collobert, C.~Fuegen, T.~Likhomanenko, G.~Synnaeve, A.~Joulin, A.~Mohamed, and E.~Dupoux, ``{Libri-Light}: A benchmark for {ASR} with limited or no supervision,'' in \emph{ICASSP}, 2020.

\bibitem{he2024emilia}
H.~He, Z.~Shang, C.~Wang, X.~Li, Y.~Gu, H.~Hua, L.~Liu, C.~Yang, J.~Li, P.~Shi \emph{et~al.}, ``Emilia: An extensive, multilingual, and diverse speech dataset for large-scale speech generation,'' \emph{arXiv preprint arXiv:2407.05361}, 2024.

\bibitem{google-usm}
Y.~Zhang, W.~Han, J.~Qin, Y.~Wang, A.~Bapna, Z.~Chen, N.~Chen, B.~Li, V.~Axelrod, G.~Wang \emph{et~al.}, ``Google {USM}: Scaling automatic speech recognition beyond 100 languages,'' \emph{arxiv:2303.01037}, 2023.

\bibitem{comanici2025gemini}
G.~Comanici, E.~Bieber, M.~Schaekermann, I.~Pasupat, N.~Sachdeva, I.~Dhillon, M.~Blistein, O.~Ram, D.~Zhang, E.~Rosen \emph{et~al.}, ``Gemini 2.5: Pushing the frontier with advanced reasoning, multimodality, long context, and next generation agentic capabilities,'' \emph{arXiv preprint arXiv:2507.06261}, 2025.

\bibitem{luger-etal-2025-building}
\BIBentryALTinterwordspacing
S.~Luger, R.~Mosquera-G{\'o}mez, A.~Mi{\l}owski, T.~Vaughan, S.~Hincapie-Monsalve, P.~Ortiz~Suarez, and K.~Bollacker, ``Building data infrastructure for low-resource languages,'' in \emph{Proceedings of the Eighth Workshop on Technologies for Machine Translation of Low-Resource Languages (LoResMT 2025)}, A.~K. Ojha, C.-h. Liu, E.~Vylomova, F.~Pirinen, J.~Washington, N.~Oco, and X.~Zhao, Eds.\hskip 1em plus 0.5em minus 0.4em\relax Albuquerque, New Mexico, U.S.A.: Association for Computational Linguistics, May 2025, pp. 154--160. [Online]. Available: \url{https://aclanthology.org/2025.loresmt-1.14/}
\BIBentrySTDinterwordspacing

\bibitem{pratap2020mls}
V.~Pratap, Q.~Xu, A.~Sriram, G.~Synnaeve, and R.~Collobert, ``{MLS}: A large-scale multilingual dataset for speech research,'' in \emph{Interspeech 2020}, pp. 2757--2761.

\bibitem{chime78}
\BIBentryALTinterwordspacing
S.~Cornell, C.~Boeddeker, T.~Park, H.~Huang, D.~Raj, M.~Wiesner, Y.~Masuyama, X.~Chang, Z.-Q. Wang, S.~Squartini, P.~Garcia, and S.~Watanabe, ``Recent trends in distant conversational speech recognition: A review of chime-7 and 8 dasr challenges,'' \emph{Computer Speech and Language}, vol.~97, p. 101901, 2026. [Online]. Available: \url{https://www.sciencedirect.com/science/article/pii/S0885230825001263}
\BIBentrySTDinterwordspacing

\bibitem{mousavi2025discrete}
\BIBentryALTinterwordspacing
P.~Mousavi, G.~Maimon, A.~Moumen, D.~Petermann, J.~Shi, H.~Wu, H.~Yang, A.~Kuznetsova, A.~Ploujnikov, R.~Marxer, B.~Ramabhadran, B.~Elizalde, L.~Lugosch, J.~Li, C.~Subakan, P.~Woodland, M.~Kim, H.~yi~Lee, S.~Watanabe, Y.~Adi, and M.~Ravanelli, ``Discrete audio tokens: More than a survey!'' \emph{Transactions on Machine Learning Research}, 2025. [Online]. Available: \url{https://openreview.net/forum?id=eqNchtvc6v}
\BIBentrySTDinterwordspacing

\bibitem{speakeasy}
\BIBentryALTinterwordspacing
S.~Brade, S.~Anderson, R.~Kumar, Z.~Jin, and A.~Truong, ``Speakeasy: Enhancing text-to-speech interactions for expressive content creation,'' in \emph{Proceedings of the 2025 CHI Conference on Human Factors in Computing Systems}, ser. CHI '25.\hskip 1em plus 0.5em minus 0.4em\relax New York, NY, USA: Association for Computing Machinery, 2025. [Online]. Available: \url{https://doi.org/10.1145/3706598.3714263}
\BIBentrySTDinterwordspacing

\bibitem{zheng2024bat}
\BIBentryALTinterwordspacing
Z.~Zheng, P.~Peng, Z.~Ma, X.~Chen, E.~Choi, and D.~Harwath, ``{BAT}: Learning to reason about spatial sounds with large language models,'' in \emph{Forty-first International Conference on Machine Learning}, 2024. [Online]. Available: \url{https://openreview.net/forum?id=kao5hRX9YA}
\BIBentrySTDinterwordspacing

\bibitem{li2025less}
C.~Li, W.~Zhang, W.~Wang, R.~Scheibler, K.~Saijo, S.~Cornell, Y.~Fu, M.~Sach, Z.~Ni, A.~Kumar \emph{et~al.}, ``Less is more: Data curation matters in scaling speech enhancement,'' \emph{arXiv preprint arXiv:2506.23859}, 2025.

\bibitem{babu2021xls}
A.~Babu, C.~Wang, A.~Tjandra, K.~Lakhotia, Q.~Xu, N.~Goyal, K.~Singh, P.~{von Platen}, Y.~Saraf, J.~Pino, A.~Baevski, A.~Conneau, and M.~Auli, ``{XLS-R: Self-supervised Cross-lingual Speech Representation Learning at Scale},'' in \emph{Interspeech 2022}, 2022, pp. 2278--2282.

\bibitem{boito2024mhubert}
M.~Z. Boito, V.~Iyer, N.~Lagos, L.~Besacier, and I.~Calapodescu, ``mhubert-147: A compact multilingual hubert model,'' \emph{arXiv preprint arXiv:2406.06371}, 2024.

\bibitem{peng25c_interspeech}
Y.~Peng, M.~Shakeel, Y.~Sudo, W.~Chen, J.~Tian, C.-J. Lin, and S.~Watanabe, ``{OWSM v4: Improving Open Whisper-Style Speech Models via Data Scaling and Cleaning},'' in \emph{{Interspeech 2025}}, 2025, pp. 2225--2229.

\bibitem{sekoyan2025canary}
M.~Sekoyan, N.~R. Koluguri, N.~Tadevosyan, P.~Zelasko, T.~Bartley, N.~Karpov, J.~Balam, and B.~Ginsburg, ``Canary-1b-v2 \& parakeet-tdt-0.6 b-v3: Efficient and high-performance models for multilingual asr and ast,'' \emph{arXiv preprint arXiv:2509.14128}, 2025.

\bibitem{commonvoice}
R.~Ardila, M.~Branson, K.~Davis, M.~Kohler, J.~Meyer, M.~Henretty, R.~Morais, L.~Saunders, F.~Tyers, and G.~Weber, ``Common voice: A massively-multilingual speech corpus,'' in \emph{LREC 2020}, 2020, pp. 4218--4222.

\bibitem{FLEURS}
A.~Conneau \emph{et~al.}, ``{FLEURS}: Few-shot learning evaluation of universal representations of speech,'' in \emph{SLT 2022}, 2022.

\bibitem{voxlingua}
J.~Valk and T.~Alumäe, ``{VOXLINGUA107}: A dataset for spoken language recognition,'' in \emph{SLT 2021}, 2021.

\bibitem{kudo-2018-subword}
T.~Kudo, ``Subword regularization: Improving neural network translation models with multiple subword candidates,'' in \emph{Proceedings of the 56th Annual Meeting of the Association for Computational Linguistics (Volume 1: Long Papers)}, Melbourne, Australia, Jul. 2018, pp. 66--75.

\bibitem{cornell23_chime}
S.~Cornell, M.~S. Wiesner, S.~Watanabe, D.~Raj, X.~Chang, P.~Garcia, Y.~Masuyam, Z.-Q. Wang, S.~Squartini, and S.~Khudanpur, ``The chime-7 dasr challenge: Distant meeting transcription with multiple devices in diverse scenarios,'' in \emph{7th International Workshop on Speech Processing in Everyday Environments (CHiME 2023)}, 2023, pp. 1--6.

\bibitem{zhang25j_interspeech}
W.~Zhang, K.~Saijo, S.~Cornell, R.~Scheibler, C.~Li, Z.~Ni, A.~Kumar, M.~Sach, W.~Wang, Y.~Fu, S.~Watanabe, T.~Fingscheidt, and Y.~Qian, ``{Lessons Learned from the URGENT 2024 Speech Enhancement Challenge},'' in \emph{{Interspeech 2025}}, 2025, pp. 853--857.

\bibitem{zen2019libritts}
H.~Zen, V.~Dang, R.~Clark, Y.~Zhang, R.~J. Weiss, Y.~Jia, Z.~Chen, and Y.~Wu, ``Libritts: A corpus derived from librispeech for text-to-speech,'' in \emph{Proc. Interspeech}, 2019, pp. 1526--1530.

\bibitem{espnet}
S.~Watanabe, T.~Hori, S.~Karita, T.~Hayashi, J.~Nishitoba, Y.~Unno, N.~{Enrique Yalta Soplin}, J.~Heymann, M.~Wiesner, N.~Chen, A.~Renduchintala, and T.~Ochiai, ``{ESP}net: End-to-end speech processing toolkit,'' in \emph{Interspeech 2018}, 2018.

\bibitem{kumar2023high}
R.~Kumar, P.~Seetharaman, A.~Luebs, I.~Kumar, and K.~Kumar, ``High-fidelity audio compression with improved rvqgan,'' \emph{Advances in Neural Information Processing Systems}, vol.~36, pp. 27\,980--27\,993, 2023.

\bibitem{espnet_codec}
J.~Shi \emph{et~al.}, ``Espnet-codec: Comprehensive training and evaluation of neural codecs for audio, music, and speech,'' \emph{arXiv preprint arXiv:2409.15897}, 2024.

\bibitem{libritts}
H.~Zen, V.~Dang, R.~Clark, Y.~Zhang, R.~J. Weiss, Y.~Jia, Z.~Chen, and Y.~Wu, ``{LibriTTS: A Corpus Derived from LibriSpeech for Text-to-Speech},'' in \emph{{Interspeech 2019}}, 2019, pp. 1526--1530.

\bibitem{shi2024versa}
J.~Shi, H.-j. Shim, J.~Tian, S.~Arora, H.~Wu, D.~Petermann, J.~Q. Yip, Y.~Zhang, Y.~Tang, W.~Zhang \emph{et~al.}, ``Versa: A versatile evaluation toolkit for speech, audio, and music,'' \emph{arXiv preprint arXiv:2412.17667}, 2024.

\bibitem{stoi}
C.~H. Taal, R.~C. Hendriks, R.~Heusdens, and J.~Jensen, ``A short-time objective intelligibility measure for time-frequency weighted noisy speech,'' in \emph{2010 IEEE International Conference on Acoustics, Speech and Signal Processing}, 2010, pp. 4214--4217.

\bibitem{ChenWavLm}
S.~Chen, C.~Wang, Z.~Chen, Y.~Wu, S.~Liu, Z.~Chen, J.~Li, N.~Kanda, T.~Yoshioka, X.~Xiao, J.~Wu, L.~Zhou, S.~Ren, Y.~Qian, Y.~Qian, J.~Wu, M.~Zeng, X.~Yu, and F.~Wei, ``{WavLM}: Large-scale self-supervised pre-training for full stack speech processing,'' \emph{IEEE JSTSP}, 2022.

\end{thebibliography}

\end{document}